# Beyond the Hype: Assessing the Performance, Trustworthiness, and Clinical Suitability of GPT3.5

Salmonn Talebi[1], Elizabeth Tong[2*] & Mohammad R. K. Mofrad[1*]

[1] Departments of Bioengineering and Mechanical Engineering, University of California, Berkeley, California, USA
[2] Department of Radiology, Stanford School of Medicine, Stanford University, Stanford, California, USA

*Co-corresponding Author:
Elizabeth Tong
Phone: (650) 723-8463
Fax: (650) 723-1909
E-mail: etong@stanford.edu
Address: 725 Welch Rd Rm 1860 Palo Alto, CA 94304

*Co-corresponding Author:
Mohammad R. K. Mofrad
Phone: (510) 643-8165
Fax: (510) 642-5835
E-mail: mofrad@berkeley.edu
Address: 208A Stanley Hall #1762, University of California, Berkeley, CA 94720-1762

**ABSTRACT**
The use of large language models (LLMs) in healthcare is gaining popularity, but their practicality and safety in clinical settings have not been thoroughly assessed. In high-stakes environments like medical settings, trust and safety are critical issues for LLMs. To address these concerns, we present an approach to evaluate the performance and trustworthiness of a GPT3.5 model for medical image protocol assignment. We compare it with a fine-tuned BERT model and a radiologist. In addition, we have a radiologist review the GPT3.5 output to evaluate its decision-making process. Our evaluation dataset consists of 4,700 physician entries across 11 imaging protocol classes spanning the entire head. Our findings suggest that the GPT3.5 performance falls behind BERT and a radiologist. However, GPT3.5 outperforms BERT in its ability to explain its decision, detect relevant word indicators, and model calibration. Furthermore, by analyzing GPT3.5's explanations for misclassifications, we reveal systematic errors that need to be resolved to enhance its safety and suitability for clinical use.

# INTRODUCTION

In recent years, the combination of algorithmic developments, increased data accessibility, and convenient computing resources has facilitated widespread adoption of unsupervised large language models (LLMs)[1]. These LLMs have showcased remarkable performance in natural language processing tasks such as text generation[2], summarization[3], and translation[4], as well as non-linguistic tasks like protein folding[5] and protein generation[6]. Prior research has demonstrated the utility of LLMs in medical contexts, where their accuracy and decision-making processes is comparable to a trained physician [7]. However, these LLMs often necessitate thousands of human-annotated labels for fine-tuning, which can be both time-consuming and expensive.

Interestingly, scaling LLMs has led to their ability to perform a wide array of tasks without requiring task-specific fine-tuning[8]. This has removed the need to create large human-annotated datasets for finetuning. Specifically, GPT3.5 represents a significant advancement over its predecessors. Developed by OpenAI, this model is built upon the same principles as earlier versions, but with notable improvements in its architecture, training techniques, and dataset size. These enhancements enable GPT3.5 to perform a wide range of tasks with minimal task-specific fine-tuning, relying solely on prompting to generate relevant responses. Recent literature has showcased the model's remarkable capabilities in various applications, including medical licensing exams and law exams[9,10]. However, there is also a growing body of research that highlights concerns regarding its limitations, such as dataset contamination, the potential for biased outputs, and difficulties in interpreting the model's decision-making process[11]. Thus, while GPT3.5 holds promise for clinical applications, it is crucial to further explore its utility and potential drawbacks in real-world healthcare settings.

In order to evaluate GPT3.5's utility in real-world clinical settings, we have annotated a custom dataset comprising clinical notes for medical imaging protocol assignment. Medical imaging is a vital component of contemporary healthcare, enabling physicians to visualize the interior organs to diagnose and manage various conditions. Clinicians frequently order radiologic studies, such as magnetic resonance imaging (MRI) or computed tomography (CT), to address clinical questions and guide treatment decisions[12].

When a physician orders an imaging study, they typically provide a concise reason for the exam, which may include the patient's signs and symptoms, brief medical history, and any relevant clinical findings. Radiologists then review these orders and recommend a radiologic protocol that best addresses the clinical question. A radiologic protocol contains a specific set of instructions that define the type of radiologic exam, the specific body part to image. The protocol is chosen by taking into consideration the reason for the exam, the patient's presentation, and expected imaging findings. In an MRI exam, the protocol may involve different imaging sequences, contrast agents, imaging planes, and field of view.

Assigning the appropriate protocol necessitates an in-depth understanding of various pathologies' radiological appearances, as well as a comprehensive knowledge of the patient's clinical presentation and medical history. It also requires familiarity with the institution's available protocols, as different facilities may possess varying capabilities and resources. In MRI, accurate protocol assignment is particularly crucial, as the chosen protocol impacts the quality and diagnostic accuracy of the exam[13, 14].

We evaluated GPT3.5 using 4,800 archived physician orders to comprehend the medical language employed to describe a given radiological exam. The novelty of this dataset, comprised of real-world clinical data not publicly available during GPT's training, offers valuable insights into GPT3.5's performance in real-world clinical applications. Additionally, we delve into the decision-making process to ascertain if GPT3.5 possesses a comprehensive understanding of various pathologies, radiological appearances, and language within the context of human anatomy and physiology.

This study offers several significant contributions, including:
- A comprehensive evaluation of GPT3.5 using a medical dataset of medical imaging protocol text, compared to existing state-of-the-art models such as BERT.
- A comparative analysis of the top word indicators for specific protocols between GPT3.5, BERT, and a radiologist.
- A demonstration of the model's ability to make complex decisions in a manner similar to that of a radiologist.
- An in-depth analysis of the model's mistakes through the elicitation of explanations of its decision-making process, highlighting potential systematic errors that may pose safety risks in clinical settings.
- The use of log probabilities to measure the model's confidence and calibration for this task across the 11 classes.

Taken together, these contributions provide important insights into the performance and limitations of GPT3.5 in a medical context, with implications for the safe deployment of large language models in healthcare settings.

## DATA

In order to evaluate a GPT3.5 for medical text classification, we have compiled a large-scale dataset for image protocol review. This dataset consists of deidentified order entries and assigned protocols for magnetic resonance (MR) neuroradiology studies that were conducted at our institution between June 2018 and July 2021. Each row in the dataset represents a single radiology order and includes the 'reason for exam', patient age and gender, and the protocol assigned by the radiologist. To ensure the accuracy and quality of the data, we performed a thorough review by an experienced radiologist (ET) with 10 years of experience. The final dataset includes 4730 recorded notes with expert-annotated imaging protocols (table 1).

Eleven protocols very commonly used at our institution were included in this database. The 'MR brain routine' protocol is the default protocol with basic MR sequences. Six out of ten protocols are optimized for specific diseases or conditions - *'MR brain demyelinating*', '*MR brain mass/metastases(mets)/infection'*, '*MR moya-moya with Diamox (mDiamox)*', '*MR seizure*', '*MR stroke*', and '*MR vascaular malformation/hemorrhage/trauma*'. There are four protocols - '*MR skull base*', '*MR sella*', '*MR nasopharynx-oropharynx*', and '*MR orbit-sinus-face*', that are optimized to image specific regions of the head.

# METHODS

This retrospective study was conducted with the approval of the Stanford Institutional Review Board (IRB) and under a waiver of informed consent. The study was approved for collaboration between Stanford University and the University of California, Berkeley.

### GPT3.5 evaluation

We approach the problem of text classification as predicting the class that corresponds to a given input text. In our dataset, we have 11 possible classes that can be predicted. To achieve this, we create a prompt which briefly explains the task at hand and limits the prediction output to the 11 target imaging protocols.

**Prompt:** "*You are being evaluated on how well you can perform radiological protocol assignment for MR imaging. Suppose we have 11 possible MR Imaging protocols: MR NASOPHARYNX OROPHARYNX, MR BRAIN MASS/METS/INFECT, MR STROKE, MR SELLA, MR BRAIN SEIZURE, MR BRAIN DEMYELINATING, MR BRAIN MOYA-MOYA DIAMOX, MR SKULL BASE, MR VASCULAR MALFORMATION/ICH/TRAUMA, MR ORBIT SINUS FACE, MR BRAIN ROUTINE.*

*We will provide you with a physician's entry and request that you identify which imaging protocol, out of the 11 options listed, should be used and explain the reasoning behind your choice.* "

The prompt combined with each physician entry was submitted via open AI's API for bulk processing. GPT3.5 responses were parsed for the predicted protocol along with its explanation. The predicted protocol was compared against our ground truth to measure the precision, recall and accuracy for each class.

### Model Baseline

In order to establish a baseline and compare the performance of GPT3.5 against finetuned large language models, we conducted experiments using several well-known pretrained models, namely BERT, BioBERT and RoBERTa. These algorithms have been used in previous studies for medical imaging protocol assignment and provide a benchmark to evaluate the effectiveness of our approach. To achieve this, we fine-tuned a pre-trained BERT, BioBERT, RoBERTa models using the HuggingFace Transformers library [28].

### Word Importance

Since GPT3.5 is a generative model, we can prompt it to explain its decision-making process and review its reasoning for systematic errors. In addition to the model's explanations, we perform an additional prompt asking for the model to output the top 3 most important words (in descending order) that it used to make its prediction. To achieve this, we include an additional request to the prompt.

**Prompt:** "*…We will provide you with a physician's entry and request that you identify which imaging protocol, out of the 11 options listed, should be used and explain the reasoning behind your choice. In addition, can you list the 3 most important words (single words) which had the greatest impact on your decision?*"

We employ the concept of "word importance" as an additional means of interpreting the model. Word importance quantifies the contribution that each word in the input text makes to the model's prediction.

In order to baseline against BERT's ability to determine relevant words that are indicative of the target protocol we use a method called integrated gradients. Integrated gradients exploit the gradient information of the model by integrating first-order derivatives. This method does not require the model to be differentiable or smooth, making it particularly suitable for large and complex models such as Transformers. We use integrated gradients to accurately estimate the importance of individual words within an input sentence.

The integrated gradients method can be formally defined as follows: let x be the input sentence, represented as a $(x1, ..., xm)$, and let $x'$ be a "blank" baseline input. We have a trained model $F$, and $F(x)n$ is the output of the model at time step n. The contribution of the $mth$ word in $x$ to the prediction of $F(x)n$ can be calculated by taking the integral of gradients along the straight line path from $x'$ to the input $x$. In other words, we are measuring how much the prediction at time step $n$ changes as we move from the baseline input $x'$ to the actual input $x$, and specifically how much the $mth$ word in $x$ contributes to this change.

The word importance value of each word in the input is calculated by summing the scalar attributions across the dimensions of the input embeddings. A positive attribution value indicates that the word contributed to the prediction made by the model, while a negative attribution value indicates that it opposed the prediction. In cases of the BERT model, which uses sub-word tokenization to divide rare words into smaller pieces, we can obtain word-level attributions that are more understandable to humans by taking the sub-word with the highest absolute attribution value as the attribution for the entire word.

**Aggregating word attribution**

To aggregate the word attribution values, we adopted a multi-text approach for each imaging protocol. Firstly, we requested the GPT3.5 model to output the top three most important words for each entry. Subsequently, we assigned a weight of 1 to the top word, 0.66 to the second most important word, 0.33 to the third most important word, and 0 to all other words not listed. The GPT3.5 word importance score was then determined by averaging the word scores across a single class for each word.

Similarly, a radiologist assigned a measure of word importance across all texts for a given imaging protocol, based on a numerical score, where a value of 1 indicates a strong influence on the radiologist's decision, 0.5 indicates a slight influence, and 0 indicates a neutral influence. The human word importance score was determined as the average of all word scores across a single imaging protocol class for each word.

In addition, Integrated Gradient was used to assign attribution scores for BERT predictions. We calculated the top aggregate words for each imaging protocol by taking the average attribution value for each word across all texts for a given imaging protocol and selecting the top words with the highest average attribution value.

To ensure the reliability of our results, we filtered out words that appeared in less than five texts for all three classification methods. These approaches were utilized to generate the lists of words the GPT3.5 model, the BERT model, and the radiologist found most important. The lists generated by the models were compared with the radiologist's lists to get the consensus of the five most important words.

**Model Calibration**

Model calibration is essential in understanding the degree of confidence in a model's prediction. To assess the calibration of GPT3.5 for this task, we prompted the model to output the log probabilities for each of the entries. These log probabilities were then transformed into a confidence value for the predicted imaging protocol, and confidence values were determined for all entries in our evaluation set.

To evaluate model calibration, we divided all entries in our evaluation set into bins with a range of [0, 1] and intervals of 0.1. For each bin, we calculated the accuracy to determine how well-calibrated our model is.

Similarly, we performed the same procedure for the BERT model. In addition, we were able to adjust the model calibration using temperature for our BERT model. Temperature scaling is a simple post-processing technique that can be used to adjust the confidence of a model's predictions. The idea behind temperature scaling is to divide the log probabilities of a model's output by a temperature parameter. This rescales the log probabilities of being more or less confident depending on the chosen temperature value. This method can help to better align the model's predicted probabilities with its true accuracy, resulting in more reliable and accurate predictions.

## RESULTS:

Our findings revealed that the GPT3.5 model had a weighted average F1 score of 0.72, which represents a significant decrease compared to the results obtained with fine-tuned large language models. In comparison, the weighted average F1 scores for BERT, BioBERT, and RoBERTa on our dataset were 0.89, 0.92, and 0.90, respectively. These results demonstrate that the performance of GPT3.5 falls behind that of state-of-the-art models that have been specifically fine-tuned for this task. Despite not being the top-performing model in certain tasks, the GPT-3.5 language model is still remarkable given that it has not been fine-tuned for these specific tasks.

**Word Importance**

We aggregated word attribution scores for each image protocol and investigated the difference in the word importance ranks between GPT3.5, BERT, and the radiologist (see Figure 1). There was good consensus among the radiologist and the models for the brain mass workup. Meningioma, the most common type of brain tumor, and lung cancer, the most common cause of brain metastases, were among the top 5 most important words. "Mets" is a shorthand term commonly used for metastases. The radiologist also prioritized words suggesting a history of treatment for brain tumors, such as 'cyberknife', and other brain tumor types. GPT3.5 selected 'headache', which is a symptom commonly associated with brain tumor. BERT, on the other hand, demonstrated a tendency to select words likely influenced by frequency, such as 'Rule' and 'date'. Similar trends are noted in the 'MR brain seizure' and 'MR stroke' protocols. Overall, the top five words selected by GPT3.5 aligned more closely with the radiologist's choices than those selected by BERT, which showed a bias towards non-specific words.

Further, we scrutinize individual texts and the model's explanations to evaluate the model's comprehension of language within the context of human anatomy and pathology. Figure 4 displays a physician's text in conjunction with the model's corresponding explanation. For BERT, we utilized word attribution scores, while GPT3.5 can directly elucidate its decision-making process. In the first example (Figure 4, top row), the indication was ' *62 year old famle with history of breast cancer, new symptoms of left headache, dizziness, left facila/upper arm numbness, and deviating to the left with walking*'. BERT prioritized the patient's history of breast cancer and headache. In elderly patients, headaches often signal the existence of a brain tumor, and cancer can metastasize from the breast to the brain. GPT3.5 correctly pinpointed headache, dizziness, and numbness in the left face and upper arm as symptoms suggesting a brain mass. While both model predicted the correct *'MR brain metastases'* protocol, GPT3.5 demonstrated a broader analysis of the complete text, whereas BERT concentrated on a select few keywords.

In the second example, (Figure 4 bottom row), the indication was '*59 year old with left posterior headache possible seizure, concern for edema on computer tomography. Brain tumor at age 18. Epilepsy with seizure and possible edema on computer tomography'*. BERT recommended 'MR brain seizure' protocol, which was incorrect, while GPT accurately suggested the 'MR brain metastases' protocol. BERT honed into the words 'seizure' and 'epilepsy' intensely, overlooking the fact that the mention of potential edema in a computerized tomography scan and the prior history of brain tumor, which collectively suggested the high possibility of a brain tumor. GPT3.5 picked up the patient's history of a brain tumor, coupled with the suspicious edema, to suggest the appropriate *'MR brain metastases'* protocol. GPT3.5 asserted that "*while an MRI to diagnose a*

*brain seizure is plausible*", the additional information compelled it to choose a different protocol to better evaluate brain metastases, mass or infections.

**Model Calibration**
Model calibration is an important consideration in machine learning, as it refers to the ability of a model to accurately estimate the probability of its predictions. In the context of natural language processing, calibration is particularly important because it helps users to understand how much they can trust the model's predictions. In comparing the performance of GPT-3.5 and BERT, on this task, it was found that GPT-3.5 was surprisingly well calibrated out of the box, while BERT was not (Figure 3). However, we were able to improve the calibration of the BERT model through temperature scaling. After temperature scaling, the calibration of the BERT model was brought in line with that of the GPT-3.5 model.

The fact that GPT-3.5 appears to be well calibrated out of the box is a significant advantage, as it means that users can have more confidence in the accuracy of its predictions without the need for extensive recalibration. This is particularly useful in applications where the reliability of the model's predictions is critical.

**Error Analysis**
In order to understand the errors made by the GPT3.5 model on the test set, we conducted an analysis of the model's explanations and looked for any systematic patterns in the mistakes. Our analysis identified seven broad categories of errors related to : (1) incomplete understanding of the protocol (44%), (2) anatomy (22%), (3) incomplete understanding of medical conditions/terminology (14%), (4) misunderstanding of acronyms (5%), (5) arbitrary (5%), (6) age-related (8%), and (7) ambiguous prompt (2%).

The most common type of mistake stemmed from GPT3.5's incomplete understanding of the imaging protocol. This category involved errors where GPT3.5 lacked in-depth understanding of the 11 protocols. Without prior training, it would be difficult for it to fully distinguish the specific sequences, anatomic coverage, or imaging techniques embedded within each protocol that were necessary for accurate diagnostics. For example, to 'evaluate left vertebral artery dissection', GPT3.5 recognized that 'vertebral artery dissection falls under the category of vascular malformation'. and recommended the 'MR VASCULAR MALFORMATION/ICH/TRAUMA' protocol. The best protocol should be 'MR STROKE', as it includes all the sequences within the 'MR VASCULAR MALFORMATION/ICH/TRAUMA' protocol, with additional sequences to look for a stroke, which is a common complication from a vertebral artery dissection. Furthermore, the 'MR VASCULAR MALFORMATION/ICH/TRAUMA' protocol is optimized to evaluate the vessels within the brain, and does not include the vessels in the neck, which is where most vertebral artery dissection occurs. This issue emphasizes the need for training GPT3.5 on a dataset that explicitly details imaging protocol information, which could help improve the model's comprehension and application of the protocols.

Anatomy-related errors, the second most common category, were attributable to GPT3.5's failure to understand a specific anatomic terminology, or the anatomical relationships of the human body. These errors were notable in instances where the model failed to comprehend spatial relationships or failed to correctly link symptoms to relevant anatomical structures. For example, to image a

'olfactory grove meningioma', the model recommended 'MR BRAIN MASS/METS/INFECTION' protocol, which was reasonable since meningioma is a type of mass, however, the best protocol should be the 'MR ORBIT/SINUS/FACE' protocol which would provide an exquisite view of the olfactory grove. Such instances highlighted a critical area for improvement, which could be addressed by incorporating more training data that emphasizes anatomical relationships and spatial reasoning.

Errors in the third category were due to GPT3.5's inability to understand medical conditions or terminology. This was apparent in cases where the model misinterpreted symptoms, disease processes, or the clinical significance of certain patient histories. For instance, GPT3.5 chose MR ORBIT/SINUS/FACE for an indication for 'chronic left facial paresthesia in a 55 year old patient', where the ground-truth was 'MR skull base'. To arrive at the correct protocol, the model needs to know that the most common cause of chronic facial pain is likely due to trigeminal neuralgia, which is most commonly caused by neurovascular compression at the prepontine cistern. As such, the most appropriate protocol for this work-up is 'MR skull base'. This again highlights the necessity of including comprehensive medical and clinical data in the training set, with particular emphasis on symptomatology and disease progression, to enhance GPT3.5's understanding and application of medical conditions and terminology.

The final categories of errors stemmed from the model's misunderstanding of acronyms, arbitrary, age-related misinterpretations, and ambiguous prompts. Misinterpretation of acronyms indicates a need for enhanced acronym disambiguation in the model's training. In our study, GPT3.5 seemed to have a better grasp on medical acronyms better than BERT. Commonly used medical acronyms such as TIA (transient ischemic attach), SCC (squamous cell carcinoma), mets (metastases), etc, were correctly interpreted by GPT3.5, but were misinterpreted by BERT models. However, less common or more cryptic acronyms, such as BOT (base of tongue), TN (trigeminal neuralgia), CK (cyberknife), were missed by both GPT3.5 and BERT, resulting in erroneous predictions.

Arbitrary errors, where the model generated non-existent or irrelevant information, show the need for further refining the model's capacity to stick closely to the given prompt and the available data. For example, the model provided the following explanation for an incorrect choice of 'MR brain seizure' protocol for an indication for 'trigeminal neuralgia in a 65 year old male' - 'The protocol for MR BRAIN SEIZURE should be used because trigeminal neuralgia can be a presenting symptom of seizures or a seizure disorder.' This explanation is medically invalid.

Age-related errors, such as the model not appropriately considering the patient's age in its analysis, could be rectified by incorporating age-based reasoning in the training data. For instance, the 'MR BRAIN DEMYLEINTATION' protocol would be used for evaluating 'hand numbness' in a young patient, whereas a 'MR BRAIN STROKE' would be more appropriate for older patients.

Lastly, ambiguous prompts highlighted a need for developing strategies for handling uncertainties or ambiguities, such as the model asking clarifying questions. For example, the prompt simply stated 'restaging', which was ambiguous but implied there was a history of malignancy, so the best protocol would be 'MR brain mass/mets/infection', but the model failed to recommend a prediction and gave this explanation, 'Without additional information about the patient's medical history or presenting symptoms, it is difficult to determine which imaging protocol to use'.

# DISCUSSION

This study explores the implications of using the GPT3.5 model in a medical context, offering important insights into the practicality and safety of applying large language models (LLMs) in healthcare settings. The results demonstrate that while GPT3.5 lags behind fine-tuned models like BERT, BioBERT, and RoBERTa in performance metrics, it offers distinct advantages in terms of interpretability and model calibration.

Despite GPT3.5's lower weighted average F1 score, the model showed a stronger capability to align its top words with those of a radiologist. This tendency was more pronounced than that of BERT, which seemed more biased towards high-frequency, non-specific words. Overall, the top five words selected by GPT3.5 aligned more closely with the radiologist's choices than BERT's, which showed a bias towards non-specific words. This suggests that GPT3.5 might better capture the nuances of medical language and clinical indications, a characteristic that could prove invaluable in healthcare applications where context sensitivity is critical. From our small dataset, GPT3.5 also exhibits the ability to take medical history and other relevant information into consideration when making its decisions. This needs to be further interrogated in larger datasets.

Furthermore, GPT3.5's ability to elucidate its decision-making process in detail could bolster trust and reliability in its predictions. This is an important aspect in a medical environment where explanation of clinical decisions is critical for patient trust, and often for regulatory requirements. Moreover, the model's accurate calibration right out of the box, as compared to BERT, which required temperature scaling, provide further confidence in its predictions.

However, the systematic errors identified from our error analysis point to areas where GPT3.5 could be improved for safer clinical deployment. The most common errors included an incomplete understanding of the imaging protocol, failure to grasp anatomical relations, and difficulty with medical conditions or terminology. These shortcomings can potentially be mitigated with more fine-tuning or the addition of supplementary resources to enhance GPT3.5's understanding of the medical domain.

Despite the model's issues with misunderstanding certain clinical information, the results of our analysis show promise for the application of GPT3.5 in healthcare settings. Even though it underperformed in comparison to models that have been fine-tuned with thousands of human-curated labels, GPT3.5 offers interpretability that can be extremely valuable in clinical settings where understanding the reasoning behind decisions is critical.

# CONCLUSION

In this study, we evaluated the performance of GPT3.5 using a large, real-world clinical dataset consisting of over 4,700 entries annotated by medical experts. Our results showed that GPT3.5 underperformed when compared to fine-tuned large language models that required tens of thousands of human-curated labels, which can be a time-consuming and costly process.

However, we also found that GPT3.5 demonstrated superior ability in identifying relevant words that are highly indicative of the target protocol compared to BERT, which struggled with finding specific anatomic structures and specialized medical terms that are relevant to humans. Further analysis of the errors revealed that the model's difficulties in understanding the imaging protocol, human anatomy, and incomplete understanding of medical conditions and terminology were the primary sources of errors.

Despite underperforming compared to fine-tuned models, GPT3.5 has potential for valuable applications in healthcare. One notable advantage of GPT3.5 is its ability to provide detailed explanations for its decision-making process. This feature can help healthcare professionals better understand the model's predictions and gain confidence in its capabilities. Additionally, GPT3.5's interpretability can help identify potential systematic errors that need to be addressed in future studies, as we were able to do through analyzing the model's explanations for misclassifications.

In comparison, BERT's ability to identify word importance can be challenging to interpret and lacks the level of detail provided by GPT3.5's explanations. However, it is important to recognize that GPT3.5's explanations may lead to overconfidence in its predictions, even when they are incorrect. This highlights the importance of remaining critical and acknowledging the limitations of the model.

Further research is necessary to fully explore the capabilities and limitations of GPT3.5 in healthcare. With continued refinement, GPT3.5 has the potential to significantly improve patient care, facilitate clinical decision-making, and advance the field of natural language processing in healthcare. Overall, while GPT3.5 may have limitations, its potential for valuable applications in healthcare makes it a promising tool that warrants further investigation.

# LIMITATIONS

There are several limitations to consider in the context of this study. First, our dataset comprised of neuroradiologic orders from a single center, and thus may be limited in its representation of the racial, social, and ethnic diversity of other regions. Validation with datasets from different institutions is necessary to more accurately compare the model's performance. Additionally, we limited the number of protocols to the ten most commonly used protocols in this study, which may not fully capture the breadth of protocols used in clinical practice. The data was collected from routine clinical work, which means that protocols were assigned by multiple radiologists with varying levels of experience, potentially leading to inter-operator variability. While the evaluation dataset is relatively large at over 4.700 entries, it is possible that additional data could further improve model performance. Additionally, it is important to note that there may be significant variations in the importance of certain words when considering the perspectives of different radiologists. In this study, we were constrained to a single radiologist when evaluating word-level agreement with BERT. However, in future studies, it would be beneficial to evaluate word importance from the perspectives of a diverse

## AUTHOR CONTRIBUTIONS
ET and ST conceived of the research study. ST contributed toward the implementation and evaluation of the study. ET performed the error analysis. ET, MM, ST managed the project vision and implementation along with writing of the manuscript.

## DATA AVAILABILITY
The datasets utilized during this study are not publicly available due to reasonable privacy and security concerns. The data is not easily redistributable to researchers other than those engaged in the Institutional Review Board-approved research collaborations with Stanford University.

## COMPETING INTERESTS
The authors declare that there are no competing interests.

## AWKNOWLEDGEMENTS
We would like to thank Andrew Dickson for reviewing and providing valuable feedback to the manuscript.

# Tables

| Protocol Name | Number of entries |
|---|---|
| MR brain demyelinating | 394 |
| MR brain mass/metastases/infection | 456 |
| MR moya-moya with Diamox | 366 |
| MR nasopharynx oropharynx | 383 |
| MR orbit sinus face | 425 |
| MR seizure | 354 |
| MR sella | 520 |
| MR skull base | 442 |
| MR stroke | 496 |
| MR vascular malformation/hemorrhage/trauma | 432 |
| MR Brain Routine | 462 |
| Total | 4,730 |

**Table 1. The 11 most commonly assigned protocols and their frequencies.**

| Subcategory F1 Score | BERT | BioBERT | RoBERTa | GPT3.5 |
|---|---|---|---|---|
| MR brain demyelinating | 0.94 | **0.94** | 0.94 | 0.88 |
| MR brain mass/mets/infection | 0.84 | **0.86** | 0.85 | 0.63 |
| MR moya-moya with Diamox | 0.99 | **0.99** | 0.99 | 0.91 |
| MR nasopharynx oropharynx | 0.91 | **0.93** | 0.89 | 0.81 |
| MR orbit sinus face | 0.88 | **0.88** | 0.86 | 0.68 |
| MR seizure | **0.96** | 0.95 | 0.94 | 0.89 |
| MR sella | 0.97 | **0.97** | 0.97 | 0.93 |
| MR skull base | 0.85 | **0.91** | 0.82 | 0.41 |
| MR stroke | 0.96 | **0.96** | 0.96 | 0.73 |
| MR vascular malformation/hemorrhage/trauma | 0.88 | 0.88 | **0.89** | 0.73 |
| MR Brain Routine | 0.82 | **0.86** | 0.83 | 0.43 |
| **Weighted Average F1 score** | 0.89 | **0.91** | 0.89 | 0.70 |

**Table 2. A comparison of imaging protocol F1 scores for each model. Best F1 score in each of the 11 protocols are bolded. GPT3.5 generally underperformed the other models in all categories.**

|  | Number of entries | Number of GPT3.5 errors | Accuracy | Number of BioBERT errors | Accuracy | Number of Human errors | Accuracy |
|---|---|---|---|---|---|---|---|
| MR brain demyelinating | 394 | 85 | 0.78 | 16 | 0.96 | 13 | **0.97** |
| MR brain mass/mets/infection | 456 | 136 | 0.70 | 38 | 0.92 | 9 | **0.98** |
| MR moya-moya with Diamox | 366 | 13 | 0.96 | 4 | **0.98** | 4 | **0.98** |
| MR nasopharynx oropharynx | 383 | 95 | 0.75 | 11 | 0.97 | 5 | **0.99** |
| MR orbit sinus face | 425 | 129 | 0.70 | 26 | 0.94 | 15 | **0.96** |
| MR seizure | 354 | 52 | 0.85 | 4 | **0.99** | 4 | **0.99** |
| MR sella | 520 | 19 | 0.96 | 13 | 0.98 | 1 | **0.99** |
| MR skull base | 442 | 332 | 0.24 | 44 | 0.90 | 20 | **0.95** |
| MR stroke | 496 | 207 | 0.58 | 25 | 0.95 | 17 | **0.97** |
| MR vascular malformation/hemorrhage/trauma | 432 | 120 | 0.72 | 47 | 0.89 | 25 | **0.94** |
| MR Brain Routine | 462 | 95 | 0.80 | 47 | 0.90 | 32 | **0.95** |
| **Weighted Average** |  |  | 0.73 |  | 0.94 |  | **0.97** |

Table 3. Performance results of the GPT3.5 and BioBERT model compared with neuroradiologists. BioBERT accuracy is near human level. GPT3.5 greatly underperforms in all categories.

**Figures:**

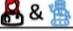

Figure 1. Top 5 words where human (trained radiologist) and GPT3.5(Top)/BERT(Bottom) agree or disagree for 3 selected protocols. Human & robot are words both human and GPT3.5/BERT agree are important. Human only are words with high human importance but low GPT3.5/BERT importance. Robot only are words with high GPT3.5/BERT importance but low human importance.

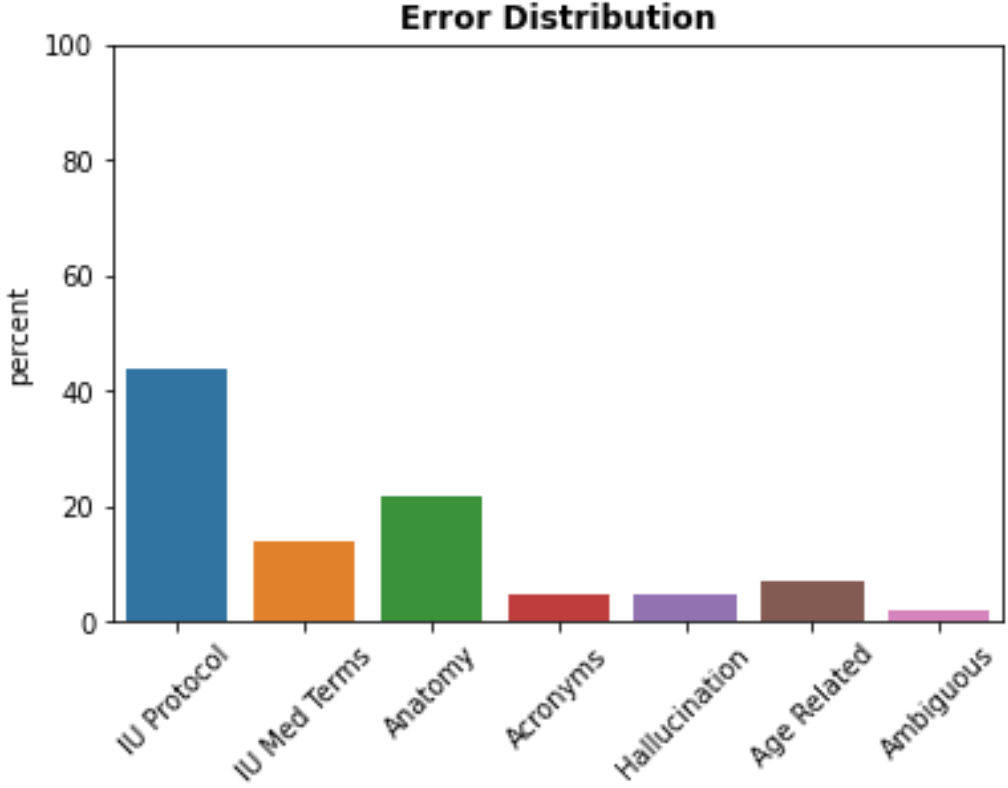

Figure 2. The bar plot decomposes the mistakes into seven categories: incomplete understanding of protocol, incomplete understanding of medical terms, anatomy related, misunderstanding of acronyms, hallucination, age related, and ambiguous errors.

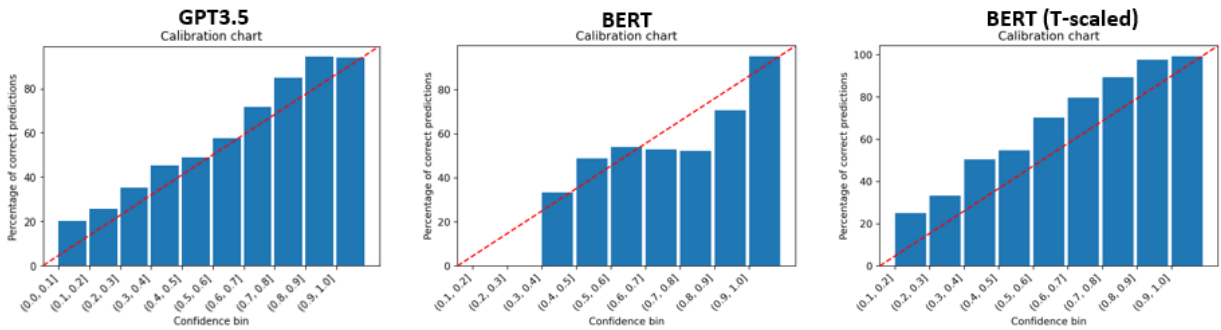

Figure 3: The figure presents a comparison of model calibrations for GPT-3.5 (left), BERT (center), and BERT with temperature scaling (right). Each subplot in the figure displays a calibration curve, illustrating the relationship between the model's predicted probabilities and the observed frequencies of correct predictions. The x-axis represents the predicted probabilities, and the y-axis represents the observed frequencies. A perfectly calibrated model would have its curve aligned with the diagonal line, indicating that the predicted probabilities match the actual outcomes.

| BERT | GPT 3.5 | True Label | BERT Explanation | GPT3.5 Explanation |
|---|---|---|---|---|
| MR Brain METS | MR Brain METS | MR Brain METS | 62yo female with history of breast cancer, new symptoms of left headache, dizziness, left facial/upper arm numbness, and deviating to the left with walking | The symptoms of headache, dizziness, and numbness in the left facial and upper arm, as well as deviation to the left while walking, are suggestive of a possible neurological deficit. These symptoms may be indicative of a brain mass, as tumors or lesions in the brain can cause neurological symptoms, depending on their location and size. In particular, the left-sided nature of these symptoms suggests that a mass or lesion in the right hemisphere of the brain may be responsible. |
| MR Brain Seizure | MR Brain METS | MR Brain METS | 59 yo w left posterior headache possible siezure, concern for edema on computer tomography. Brain tumor at age 18. epilepsy w seizure and possible edema on computer tomography. gender male | Based on the patient's history of a brain tumor at age 18 and current symptoms of left posterior headache and possible seizure, as well as the concern for edema on CT, the imaging protocol that should be used is MR BRAIN MASS/METS/INFECT. This protocol is appropriate for imaging the brain to identify any potential tumors, metastases, or infections that may be responsible for the patient's symptoms.<br><br>While MR Brain Seizure may be considered if there is suspicion of seizure activity, the patient's history of a previous brain tumor and current concern for edema make it more appropriate to use the MR BRAIN MASS/METS/INFECT protocol to evaluate for any possible pathology. |

**Figure 4:** Selected samples from the dataset. The indication for the exam is provided by the ordering physician, which briefly summarizes the symptoms, relevant medical history, and the medical questions. The 'true label' is the protocol, assigned manually by a trained radiologist, that is most suitable for the indication. The 'predicted label' is the protocol predicted by the BERT and GPT 3.5. BERT decision explanations are given by word importance (blue is more important, red is least important). GPT3.5 explanations are given by the text output of the prompt.